\documentclass[sigconf]{acmart}

\usepackage{amsmath}
\usepackage{graphicx}
\usepackage{comment}
\DeclareMathOperator*{\argmin}{argmin}

\usepackage{ stmaryrd }
\usepackage{caption}
\usepackage{todonotes}
\usepackage{subfig,float}
\usepackage{bm}
\usepackage{makecell}
\AtBeginDocument{%
  }

\copyrightyear{2023}
\acmYear{2023}
\setcopyright{acmlicensed}\acmConference[FAccT '23]{2023 ACM Conference on Fairness, Accountability, and Transparency}{June 12--15, 2023}{Chicago, IL, USA}
\acmBooktitle{2023 ACM Conference on Fairness, Accountability, and Transparency (FAccT '23), June 12--15, 2023, Chicago, IL, USA} \acmPrice{15.00}
\acmDOI{10.1145/3593013.3594122}
\acmISBN{979-8-4007-0192-4/23/06}

\usepackage{xcolor}

\begin{document}

\title{Achieving Diversity in Counterfactual Explanations:\\ a Review and Discussion}
\author{Thibault Laugel}
\authornote{Both authors contributed equally to this research.}
\orcid{0000-0002-5921-3225}
\affiliation{%
  \institution{AXA}
  \city{Paris}
  \country{France}
}
\email{thibault.laugel@axa.com}

\author{Adulam Jeyasothy}
\authornotemark[1]
\affiliation{%
  \institution{Sorbonne Université, CNRS, LIP6, }
  \city{F-75005, Paris}
  \country{France}}
\email{adulam.jeyasothy@lip6.fr}

\author{Marie-Jeanne Lesot}
\affiliation{%
  \institution{Sorbonne Université, CNRS, LIP6, }
  \city{F-75005, Paris}
  \country{France}}

\author{Christophe Marsala}
\affiliation{%
  \institution{Sorbonne Université, CNRS, LIP6,}
  \city{F-75005, Paris}
  \country{France}}

\author{Marcin Detyniecki}
\affiliation{%
  \institution{AXA, Paris, France}
  \country{}}
\affiliation{%
  \institution{Sorbonne Université, CNRS, LIP6, F-75005, Paris, France}
  \country{}}
\affiliation{%
  \institution{Polish Academy of Science, IBS PAN}
  \city{Warsaw}
  \country{Poland}}

\renewcommand{\shortauthors}{Anonymous authors}

\newcommand{\comm}[1]{} 

\begin{abstract}
In the field of Explainable Artificial Intelligence (XAI), counterfactual examples explain to a user the predictions of a trained decision model by indicating the modifications to be made to the instance so as to change its associated prediction. These counterfactual examples are generally defined as solutions to an optimization problem whose cost function combines several criteria that quantify desiderata for a good explanation meeting user needs. A large variety of such appropriate properties can be considered, as the user needs are generally unknown and 
differ from one user to another; their selection and formalization is difficult. To circumvent this issue, several approaches propose to generate, rather than a single one, a set of diverse counterfactual examples to explain a prediction. This paper proposes a review of the numerous, sometimes conflicting, definitions that have been proposed for this notion of diversity. It discusses their underlying principles as well as the hypotheses on the user needs they rely on and proposes to categorize them along several dimensions (explicit vs implicit, universe in which they are defined, level at which they apply),
leading to the identification of further research challenges on this topic.

\end{abstract}




\maketitle




\section{Introduction}

Over the last years, the need for a better understanding, and accountability, of Machine Learning systems has led to the soaring of domains around the topic of Responsible Artificial Intelligence. Among these, the eXplainable Artificial Intelligence (XAI) domain~\cite{Burkart2021,Linardatos2021} focuses on the generation of explanations for the decisions of AI and Machine Learning models. 
In particular, local post-hoc methods~\cite{guidotti2019Bboxsurvey} 
aim at generating explanations regarding the prediction performed by a given trained classifier (post-hoc property) for a given data instance of interest (local property).  They come in different formats such as feature importance (e.g LIME~\cite{ribeiro2016} and SHAP~\cite{lundberg2017}) or counterfactual examples~\cite{wachter2018} (e.g. Growing Spheres~\cite{laugel2018} and FACE~\cite{Poyiadzi2020}).

However, generating explanations has been proven to be a difficult task, due to the subjective and vague nature of the concept of interpretability~\cite{Srinivasan2020}, and thus the difficulty to define what a good explanation is. This topic has been explored from the point of view of cognitive sciences as well as educational sciences among others, as for instance summarised in the rich survey proposed by Miller~\cite{MILLER_2019} that underlines the wide range of possibly desirable properties. 
As a result, numerous explainability approaches have been proposed over the last years, mirroring the absence of consensus regarding the properties explanations should satisfy.
This issue is even more prevalent, and visible in the case of counterfactual examples~\cite{wachter2018,guidotti2022CFsurvey}. Indeed, this type of instance-based explanations relies on solving an optimization problem, and therefore explicitly depends on the selection of considered quality criteria. Although some of these seem to be consensual,
such as the closeness to the instance of interest and the sparsity of the explanation~\cite{verma2020counterfactual}, there is generally no global agreement over numerous additional possibly desirable properties, nor on how to formalize them in measurable numerical criteria.
Moreover, once selected, these criteria most often need to be  combined or aggregated to define a multi-criteria optimization problem, generally resulting in the generation a single final explanation~(see e.g. \cite{wachter2018,laugel2018,mahajan2019, Poyiadzi2020, Artelt2020}). 
Consequently, in addition to the issue of identifying the most relevant criteria assessing the quality of a candidate counterfactual example, the choice of this aggregation operator is also obviously crucial and plays a major role in the implementation of the definition of a good explanation. The possibly subjective and personal characteristic of the latter can be considered as advocating for making it dependent on the targeted user, his/her specific needs and prior knowledge. 
Yet, the selection of this aggregation is rarely motivated and often implicitly relies on the principle that all of the selected criteria should be optimized at the same time. However, this is often impossible, due to their mutual dependencies and the trade-offs existing between them.

This paper proposes to discuss the importance of this aggregation operator and shows that failing to motivate its choice, as it is often the case in the literature, may lead to unsatisfactory explanations. This provides additional  arguments to questioning the relevance of generating a single counterfactual example and arguing that the generation of multiple counterfactual examples may be more suited to meet (sometimes unformulated) user needs, a commonly identified shortcoming of interpretability. Turning to existing approaches that make it possible to generate such multiple explanations, the paper then proposes to discuss the notion of diversity they usually integrate, so that the output explanations are not redundant one with another. It offers to categorize these approaches based on the strategies they consider for this multiple explanation generation problem, and discusses how these help overcoming one pitfall of interpretability methods: matching explanations to unobserved user needs. The contributions of the paper are thus both to discuss the generally poorly covered topic of combining quality criteria for XAI methods, and to propose a review of the current literature on diverse counterfactual methods.

The paper is organized as follows. Section~\ref{sec:background} first reminds  the formal definition of explanations based on counterfactual examples as well as the most common quality criteria proposed in existing works. Section~\ref{sec:aggregation} then  discusses the importance of the operator used to aggregate these criteria and its potential overlooked undesired 
consequences. After discussing how providing multiple explanations to users might help circumventing these issues, we propose in Section~\ref{sec:diverse} a survey of different strategies to achieve this goal, discussing the various notions of diversity they rely on and how they integrate it in the explanation generation process. This study opens the way to identifying research challenges, as discussed in Section~\ref{sec:discussion}. Section~\ref{sec:conclusion} concludes the paper. 

\section{Background: Counterfactual Explanations and Common Quality Criteria}
\label{sec:background}

Within the wide domain of XAI~\cite{tjoa2020, molnar2022, Saeed2023}, counterfactual example explanations (see \cite{artelt2019survey,verma2020counterfactual,Martens2021,guidotti2022CFsurvey} for dedicated surveys  
) focus on the case where a user wants to understand the reason for a given prediction: given a data instance of interest, denoted~$x$ in the following, and a trained machine learning model, denoted~$f$, counterfactual examples aim at providing insights to understand the generated prediction~$f(x)$. More precisely, counterfactual examples aim at answering the question~"Why $f(x)$ and not $c'$?", where $c'$ denotes a possible alternative output the model~$f$ may have given.  The answer to this question is expressed through a set of modifications that can be applied to~$x$ to obtain the different prediction by the model, which amounts to answering the question: "What changes would be required to modify this prediction?". 

This type of explanations, directly meeting the desirable \emph{contrastive} property for a good explanation proposed by Miller~\cite{MILLER_2019}, has been praised for their higher transparency~\cite{wachter2018} and actionability~\cite{karimi2021survey} as compared to other types of explanations. Indeed, they have the benefit of directly describing actions that can be performed by the user to have a recourse on the prediction. On the contrary, other forms of explanations such as local feature importance vectors rely on debatable, sometimes opaque definitions of importance, and therefore have been shown to be often misunderstood (see for instance~\cite{Kaur2020} for misuses of the explanations provided by SHAP~\cite{lundberg2017}).

There exists numerous approaches to build counterfactual example explanations (e.g. see \cite{guidotti2022CFsurvey} for one of the latest surveys). This section does not aim at providing such an exhaustive overview, but rather to summarize the required background on which the later discussion is built: after providing a reminder about the general formulation problem to generate counterfactual examples, it discusses some desirable properties they have  been required to offer. 


\subsection{General Formulation of the Counterfactual Example Generation Problem}
\label{sec:background-generalCF}

This section introduces the general counterfactual example problem, focusing on the case when the machine learning model to be explained is a binary classifier, which is the most classical one. In this case, the alternative output~$c'$ given by this classifier can only be the other class. 
Denoting $\mathcal{X}$ the data feature space and $\mathcal{Y}$ the (binary) label space, the classifier is $f:\mathcal{X}\longrightarrow\mathcal{Y}$ and $x\in \mathcal{X}$ is the data instance about which the user requests an explanation regarding its associated prediction~$f(x)$.
A counterfactual example explaining this prediction is then formally defined as: 
\begin{equation}
    e^* = \argmin_{e \in \mathcal{E}} pen_x(e)
    \label{eq:generic-CF-problem}
\end{equation}
that depends on the search space~$\mathcal{E}$ and the penalty function~$pen_x$ . The former,~$\mathcal{E}$, defines the space in which the final explanation, the counterfactual example, is allowed to evolve. In its most general form, it is defined  as the set of all instances predicted to belong to a different class than~$x$, formally:
\begin{equation}
    \mathcal{E} = \{ e \in \mathcal{X}, f(e) \neq f(x)\}
    \label{eq:research_space}
\end{equation} 
This formulation assumes that all instances from the opposite class from~$x$ are equivalent, which is not always the case. For instance, some works, e.g.~\cite{wachter2018}, take into account, when available, the classification score given as output by the classifier.

The penalty function $pen_x$ to be minimized, measures the cost of a candidate explanation, which corresponds to a decreasing function of its quality. The most commonly accepted definition imposes that the counterfactual example must be as close as possible to the instance of interest, so as to minimize the amount of changes needed to alter the prediction, i.e. to lower the efforts required from the user to meet his/her objective. A common choice to capture numerically this closeness requirement is to define the penalty function as a distance between the candidate counterfactual example and the instance of interest:
\begin{equation}
    pen_x(e) = \parallel x-e \parallel
\end{equation} 

Existing works often consider $l_1$~\cite{wachter2018} or $l_2$~\cite{lash2017,laugel2018} distances. However, other distances are sometimes used, such as weighted Manhattan distance~\cite{Artelt2020}, or elastic net loss~\cite{VanLooveren2021}. 

Although alternative formulations of the counterfactual problem can be found in some works (e.g.~\cite{wachter2018}) framing it as a weighted sum of the penalty function and a classification confidence score
,
the general counterfactual problem formulation presented in Equation~\ref{eq:generic-CF-problem} can be used to understand most existing counterfactual approaches. 
However, it is important to keep in mind that a lot of approaches to generate counterfactual examples actually rely on heuristic and do not make explicit the underlying cost function they optimize. 


\subsection{Some Desirable Properties for Counterfactual Examples}
\label{sec:background-criteria}

Beside being close to the instance whose prediction is to be explained, other desirable properties have been identified for counterfactual examples. These further constrain the optimization problem through associated criteria, aiming to lead to more understandable, useful or relevant explanations. Some existing literature reviews (see e.g.~\cite{verma2020counterfactual}) have proposed to categorize counterfactual methods depending on these formulated objectives. In this section, we list some of them, depending on whether they are general for any candidate or aim at taking into account some enriched information about their context, if available. For the latter, we distinguish between data-contextualisation and user-contextualisation. Table~\ref{tab:criteria-categorization} summarises this categorisation. 
For each desirable property, we also present the associated criteria generally used to represent it in the optimization problem. 

\begin{table}[t]
    \centering
    \begin{tabular}{|l||c|c|c|}
        \toprule
        \textbf{Type} & General & Data-context. & User-context. \\
         \midrule
         \textbf{Property} & \makecell{closeness, \\ sparsity} & \makecell{local density \\ proximity \\   path density \\  justification} &  
         \makecell{actionability \\ causality, \\ personalization}\\
         \midrule
         \textbf{Dependency} &  $x$ & $x, X$ & $x, U (opt.: X)$\\
     \bottomrule
         
    \end{tabular}
    \caption{Categorization of the most frequently considered desirable properties for a candidate counterfactual example~$e$. The last row indicates the parameters they depend on: $x$ is the data instance of interest, $X$ a set of data points, $U$ the user who receives the explanation. }
    \label{tab:criteria-categorization}
\end{table}

\subsubsection{Sparsity.} One of the most frequent desiderata for counterfactual explanations is the sparsity of the explanation vector. Indeed, it ensures that the effort required to alter the prediction of $x$ focuses on a low number of features, making it more understandable and actionable by the user. Sparsity can be numerically measured by the $l_0$ distance~\cite{laugel2018,guidotti2019lore,Dandl_2020}, and optimized directly or through  the $l_1$ distance ~\cite{wachter2018,Artelt2020}. 

\subsubsection{Data-contextualisation Criteria.} Blindly minimizing a distance function has been observed to lead to unrealistic explanations: the solution to the optimisation problem may for instance lie out of the data distribution~\cite{laugel2019issues,laugel2019dangers}, making it difficult to understand or even fully absurd to the user receiving the explanation.  To avoid this issue, some constraints on where the counterfactual example should lie are often proposed and aim at providing a contextual enrichment to its generation, depending on possibly available additional information about the data. 

 These constraints can for instance be directly formulated as hard convex constraints on the allowed feature range~\cite{lash2017,karimi2020model, ramakrishnan2020synthesizing,VanLooveren2021}, or by imposing that the generated counterfactual example belongs to a dense region of the distribution~\cite{Artelt2020,Poyiadzi2020}. Rather than density constraints, Laugel et al.~\cite{laugel2019dangers} argue that counterfactual explanations should be "justified" by the ground-truth data, guaranteeing their realism by their connection to observed data points.  
On a different note, model-based counterfactual approaches, generally relying on Autoencoders Generative Adversarial networks, impose through a reconstruction loss that the generated counterfactual should be close (proximity criterion), for instance using Euclidean distance, to ground-truth instances~\cite{mahajan2019,Mothilal2020,pawelczyk2020CF,VanLooveren2021}.

Instead of solely imposing constraint on the counterfactual example to generate,
Poyiadzi et al.~\cite{Poyiadzi2020} argues that the entirety of the path connecting it to~$x$ should be in data-dense regions only, to ensure that the intermediary steps are feasible.

\subsubsection{User-centered Contextualisation Criteria.}
Measuring the quality of a candidate counterfactual example in the context of its use should also be made dependent on the user it is generated for, allowing for the generation of personalised explanations~\cite{Ustun2019,mahajan2019,Jeyasothy2022}: beyond the context provided by the domain and the other data among which it is looked for, taking into account the user context makes it possible to get back to a human-in-the-loop paradigm that is crucial in the XAI domain. 
Indeed, as mentioned in the introduction, 
actionability is often one of the strongest arguments in favor of the use of counterfactual explanations. As such, it has been included by some works as an explicit objective to guarantee more useful explanations, under the assumption that such user information is available. 

Some works~\cite{Ustun2019,karimi2020model} 
for instance consider that a set of editable features is provided by the users, so that an actionable counterfactual explanation is one that requires modification along these features only. 
On a different note, actionability is also integrated through causal constraints (see~\cite{karimi2021survey} for a survey on actionable recourse). 
The underlying assumption is that the changes along different features proposed by a counterfactual explanation are not independent: in order to be actionable, an explanation should take into account the causal relationships between features. The latter can for instance be modelled within 
Pearl's framework~\cite{pearl2009} to represent a causal graph and possibly
structural causal equations describing the causal interactions between the features. Actionability is then measured as the extent to which the explanation fits this graph (see for instance~\cite{mahajan2019,karimi2021algorithmic}, and more generally~\cite{pawelczykcarla,karimi2021survey} for recent surveys). 
Another user-centered constraint is proposed by Jeyasothy et al~\cite{Jeyasothy2022}, who argue that explanations should be personalized and adapted to the user's knowledge for it to be understood. A similar notion is proposed in~\cite{yadav2021low}, where a personalized cost function is proposed to answer user's needs. 

In addition to these properties, mostly centered around how easy to understand and use an explanation is, other criteria, out of the scope of this work, could be mentioned. These include for instance constraints that the explanation provider may want to impose, such as information leakage risk~\cite{pawelczyk2022trade} or robustness to manipulation~\cite{slack2021counterfactual}.




\section{The implicit difficulty of generating one Counterfactual Explanation}
\label{sec:aggregation}

As discussed in the previous section, the definition of interpretability objectives and numerical criteria to measure the extent to which they are achieved is a difficult and complex task. However, selecting  which properties are the most relevant for a given problem is only one of the issues to be considered: once they have been expressed, 
either by the user or the machine learning practitioner, most existing counterfactual approaches then combine them into an aggregated cost function, to be minimised to generate a unique explanation. 
Although a rich literature on aggregation operators exists (see e.g.~\cite{Mesiar02,Grabisch09,detyniecki2001fundamentals}), it is rarely leveraged in the field of XAI, leaving the topic of combining explainability criteria, to the best of our knowledge, rarely discussed.
Yet, this way of combining different criteria obviously directly impacts the generated explanation.
After detailing how this combination is usually done in the state of the art, we propose in this section a discussion to question the relevance of this step itself. 

\subsection{Criteria Combination Methods}
\label{sec:aggregation-edl}

Although the properties and their associated criteria presented in the previous section are by nature all desirable, 
it is usually impossible by design to maximize all of them. 
For instance, optimizing the sparsity of a counterfactual explanation is often at odds with maximizing its closeness to the instance of interest, 
as shown by~\cite{laugel2018} for instance. As a result, conjunctive aggregation operators, that require all criteria to be simultaneously satisfied, are  rarely used by counterfactual methods.
Recognizing this impossibility, some approaches thus define an objective function that constitutes an explicit trade-off between the various criteria. This is for instance how Mahajan et al.~\cite{mahajan2019} propose to aggregate the penalty of the explanation (measured by the $l_1$ distance) and the degree to which it satisfies the considered causal constraints (causal distance). Similarly, Jeyasothy et al.~\cite{Jeyasothy2022} balance the penalty with the incompatibility to the user knowledge using a weighted average, to propose a personalized explanation.

However, specifying the right balance between criteria can be difficult for the user. This is even truer as the various presented criteria  are often not commensurable. To circumvent this issue, instead of combining several criteria into a trade-off objective, numerous other approaches propose to impose a priority order between the criteria. By combining them into (explicit or implicit) constrained optimization problems, the objective becomes to generate the closest counterfactual example (i.e. minimizing the penalty function) in a subspace defined by constraints on other criteria. This optimization subspace may for instance be defined by constraints on density~\cite{Poyiadzi2020,Artelt2020}, or actionability~\cite{Ustun2019}. 
On the contrary, other approaches such as Growing Spheres~\cite{laugel2018} and LORE~\cite{guidotti2019lore} optimize the sparsity of the counterfactual explanation after optimizing its penalty. 

Specifying this priority order is also connected to the proposed optimization process: due to the general model-agnostic (and sometimes data-agnostic) paradigm  
considered by these approaches, imposing a priority order between criteria also helps with the optimization of the objective function. Indeed the subspace satisfying the higher-order constraint can then be identified in a preprocessing step (e.g. the construction of a graph for FACE~\cite{Poyiadzi2020} or the generation of a neighborhood for LORE~\cite{guidotti2019lore}).





\subsection{The Underestimated Consequences of the Aggregation Step}
\label{sec:aggregation-issues}

While much discussion is generally proposed to motivate the choice of the desired explanation properties, on the other hand, the definition of the aggregation operator used to combine these criteria is, to the best of our knowledge, rarely defended. 
As mentioned earlier, numerous counterfactual approaches rely on heuristics, and as such do not even have an explicit optimization problem, leaving this aggregation to be  implicit.
Yet, we show in this section that in both cases, the resulting aggregation of these criteria also has important, potentially undesirable, consequences on the final explanation. The importance of this choice is underlined in Section~\ref{sec:aggregation-importance}, while Section~\ref{sec:aggregation-undesirable} discusses the potential undersirable consequences of the aggregation.

\subsubsection{The aggregation operator: undiscussed, yet impactful}
\label{sec:aggregation-importance}
Although it may seem obvious in a multicriteria optimization perspective that the aggregation operator directly impacts the nature of the solution, to the best of our knowledge this point is rarely discussed in explanation approaches. This may seem surprising, especially as some existing approaches propose different heuristics for the same groups of criteria, therefore essentially differing from one another in terms of how these are combined.
For instance, Growing Spheres~\cite{laugel2018} and LORE~\cite{guidotti2019lore} both optimize the closeness of the explanation (measured by the Euclidean distance) and its sparsity (measured with the $l_0$ distance). 

\begin{figure}[t]%
\centering\includegraphics[width=0.45\linewidth]{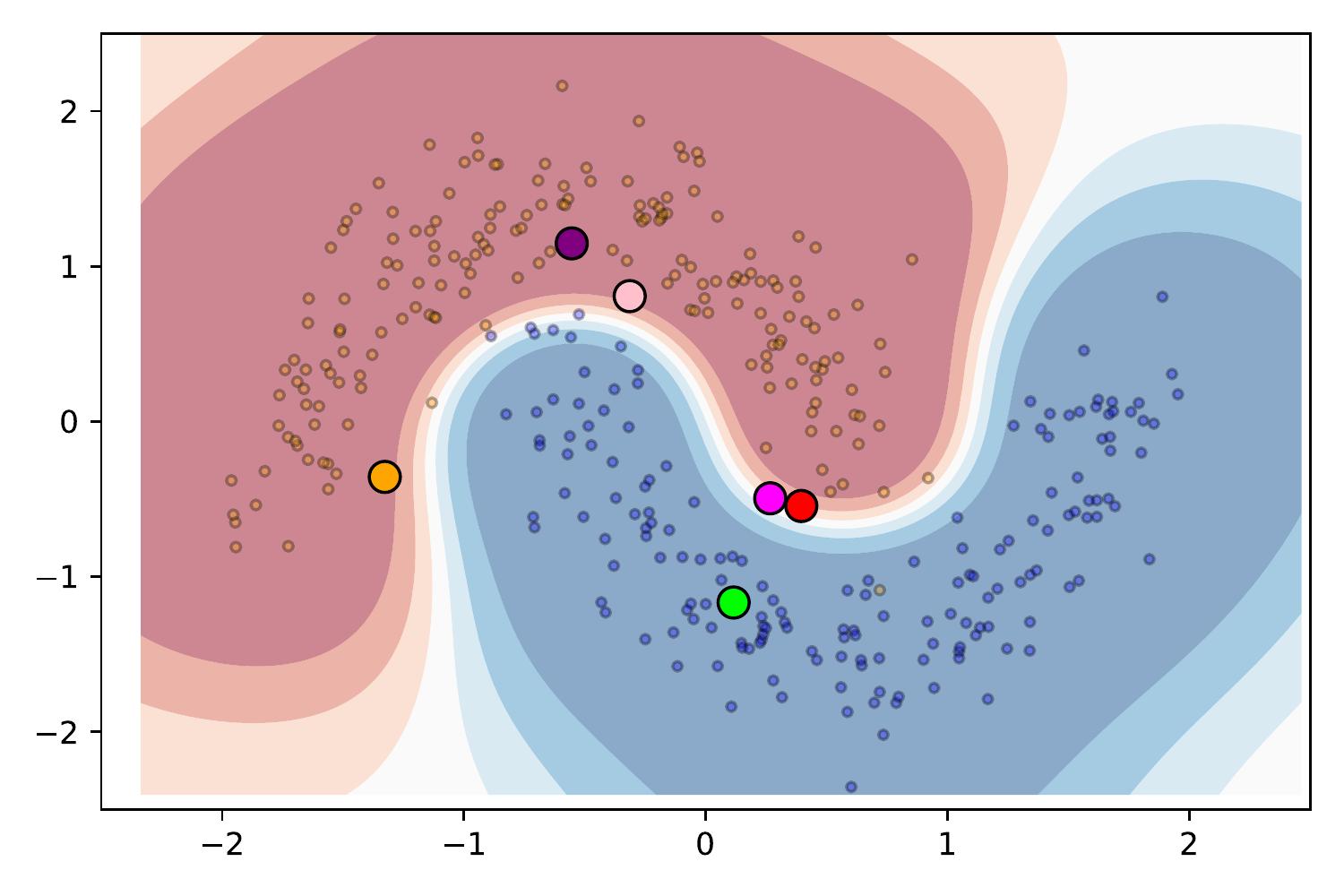}
\centering\includegraphics[width=0.45\linewidth]{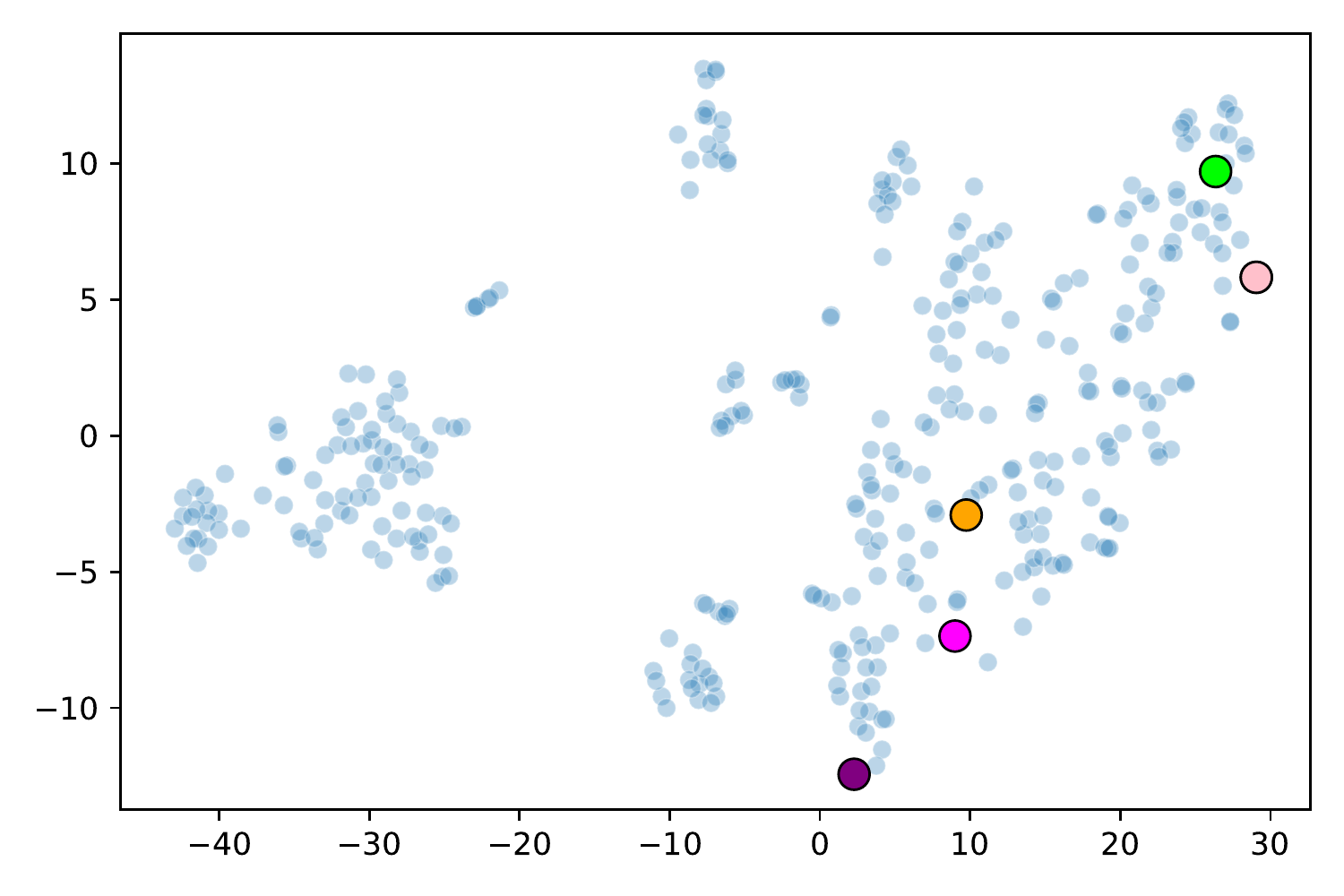}
\caption{Illustration of the aggregation operator impact: the orange, red, pink and magenta points represent the counterfactual examples generated to explain the prediction associated to the instance of interest represented by the green point, when considering several types of aggregation (see details in the text). Left: half-moons dataset. Right: 2D t-SNE projection of the Boston dataset.}
\Description{}
\label{fig:aggregation_issues}
\end{figure}

More generally, from an optimization perspective, given two criteria that cannot been optimized simultaneously (as it is often the case for counterfactual explanations), the set of possible solutions would be the Pareto front, highlighting the diversity of the possible solutions (as considered by~\cite{Dandl_2020}). To illustrate this point, we conduct a simple experiment by generating multiple counterfactual explanations optimizing the same criteria with different aggregation operators. More precisely, we consider the 2-dimensional half-moons dataset\footnote{\url{https://scikit-learn.org/stable/modules/generated/sklearn.datasets.make_moons.html}} and the Boston dataset\footnote{\url{https://www.cs.toronto.edu/~delve/data/boston/bostonDetail.html}}, on which two classifiers are trained: a SVM classifier for the half-moons dataset ($0.99$ in accuracy), and a Random Forest classifier for the Boston dataset ($0.86$ in accuracy). Figure~\ref{fig:aggregation_issues} displays illustrations of the experiment: on the left, the half-moons dataset; on the right, a 2D representation of the Boston dataset using t-SNE~\citep{van2008visualizing}.
In both cases, an instance~$x$ whose prediction is to be explained, represented by the green point, is randomly picked. We then consider two desirable properties for counterfactual examples: the closeness of the explanation, as measured by the Euclidean distance, and its belonging to a dense region of its targeted class, as measured by the log-likelihood of the counterfactual example under a Gaussian Kernel Density Estimation (KDE) trained on the corresponding data. These two criteria are then combined using various aggregation operators found in the literature, and the resulting counterfactual examples shown on the figure in different colors: a weighted sum~\cite{mahajan2019,Mothilal2020,Jeyasothy2022} (pink), the maximization of the closeness under density constraints~\cite{Poyiadzi2020} (magenta), the maximization of the density under closeness constraints~\cite{laugel2018,guidotti2019lore} (orange), and a maximisation of each criterion independently (red for the closeness, purple for the density). As expected, the resulting counterfactual examples are quite scattered across the dataset, 
covering regions of the feature space characterized by different decision boundaries. This illustrates the importance of the aggregation operator for counterfactual explanations. 

\subsubsection{Undesirable consequences}
\label{sec:aggregation-undesirable}

Beside potentially leading to drastically different results, the proposed choice of the aggregation operators sometimes raises questions in terms of relevance and user needs. For instance, optimizing for criteria which are mathematically at odds with one another may lead to trade-off solutions, that satisfy a bit of both, without satisfying fully any of them. 
This can be seen as especially problematic in the case when the properties involved belong to different property categories (see. Section~\ref{sec:background-criteria}), as they involve parameters that can hardly be compared (e.g. data vs. user preferences).
It would then seem unlikely that users with non-technical background, who are often considered to be potential users of XAI methods, would understand that generated explanation does not satisfy all of the desired properties. For instance in the case of an aggregation performed with a weighted sum~\cite{mahajan2019}, users may not understand that 
the \emph{causality} of the explanation is not guaranteed, let alone that it would come at the expense of more easiness of action (closeness of the explanation). 
The high number of possible desirable explanation objectives makes this issue all the more problematic.

As a conclusion, the criteria combination should be discussed just as much as which their individual selection. 
Considering the difficulty of aggregating numerous properties, forcing it through heuristics thus seems questionable. On the other hand, the variety of possible criteria, especially the ones where there is a tradeoff involved, pushes towards the generation of multiple explanations, that could then for instance focus on different properties.


\section{Generating Diverse Explanations}
\label{sec:diverse}

Contrary to the approaches discussed in the previous section, some other methods aim to provide users with multiple explanations at the same time to explain a single prediction. In this section, after providing some additional arguments for these approaches, we categorize existing works based on the diversity function they use to generate several explanations. We end the section with a discussion over these notions of diversity, showing how they help addressing some of the identified shortcomings of explainability approaches.

\subsection{Additional Motivations and Discussion}
\label{sec:diverse-motivations}

Although a portion of existing works proposing diverse counterfactuals state some motivations for doing so, these efforts remain generally light and scattered. We seek in this section to provide a stronger and more in-depth case for multiple explanations, centered around two arguments that complement the one previously developed in Section~\ref{sec:aggregation-issues}: (i) results from social and cognitive sciences have proven the strong benefits of using multiple explanations to teach complex concepts; (ii) having multiple explanations can help overcoming one of the main shortcomings of machine learning interpretability, namely identifying and addressing unformulated user needs.

\subsubsection{Insights from social sciences: More (carefully selected) explanations leads to better understanding.} 
In various scientific fields, providing multiple explanations has long been identified as a key factor for a better understanding on complex concepts. For instance in a clinical context, Wang et al. \cite{wang2019designing} insist that multiple explanations are required  for physicians to make better diagnostics.
On the same note, stronger conclusions are drawn in the fields of education and psychology: when using analogies to teach complex concepts to medical students, providing a single explanation was shown to create a high risk of misconceptions~\cite{spiro1989multiple}. On the other hand, providing multiple, carefully selected, analogies is presented as requirement for a good understanding. 
More generally, gathering and discussing some insights from several social sciences, Miller~\cite{MILLER_2019} insists that causes for an event must be seen as multiple, and that one important aspect of generating a good explanation is the selection by the user of his/her preferred explanation among a set of plausible ones.  
More recently, 
Bove et al.~\cite{claraPREPRINT} empirically show the benefits of providing multiple explanations to users on a classification task: the latter leads to 
an increase both in terms of objective comprehension and subjective satisfaction.

\subsubsection{Multiple explanations may help in overcoming a critical issue of interpretability: matching (unknown) user needs.}

One of the most crucial identified pain point of interpretability in general is the difficulty to determine user needs. Although it is commonly recognized that explanations should be adapted to those needs, as well as to the user's characteristics such as their knowledge and expertise (see e.g.~\cite{brod2013influence} for explanations in general, and~\cite{Gentile2021} for explanations for machine learning models), providing generic tools to do so is complex and constitutes a  poorly covered  task (some exceptions include for instance~\cite{doshivelez2017,vermeire2022choose}).
In this regard, it seems illusory to hope for one explanation method to satisfy these undeclared user needs, especially as humans have been known to perceive feature interactions and effects differently~\cite{grgic2018}.
Generating several explanations and letting the user choose the most relevant one(s) to them seems, in this vein, a way to leave this 'extra-mile' task of mapping user needs to explanations to the user, i.e. have the user select the most suited explanation and discount others~\cite{hilton1996discounting,MILLER_2019,sullivan2022explanation}.

Although some of these reasons have also been identified by previous machine learning works, there still are few works in this direction. In the next section, we present them and discuss \emph{how} they propose multiple explanations.




\subsection{Existing Diverse Counterfactual Explanation Approaches}
\label{sec:diverse-edl}

The arguments presented in Sections~\ref{sec:aggregation-issues} and~\ref{sec:diverse-motivations} have led several approaches to explain predictions through multiple counterfactual examples. For this purpose, most of them rely on the notion of \emph{diversity} (see e.g.~\cite{russell2019efficient,Mothilal2020}), imposing that the multiple explanations differ from one another, to avoid redundancies and "propose various alternatives when user preferences are not known". Yet, mirroring the lack of consensus among desirable properties for explanations, this notion of diversity has been defined in various ways, generally with few discussions associated. In this section, we review these notions of diversity, discussing the existing literature of diverse counterfactual examples. These discussions are summarized in Table~\ref{tab:diverse-survey}. The first three subsections detail in turn the three types of diversity we propose to distinguish, respectively named criteria, feature space and actions. Other criteria, related to the optimization procedure itself, are discussed in the fourth subsection.

\begin{table*}[t]
    \centering
    \resizebox{\textwidth}{!}
    {\begin{tabular}{l|ll|lcc}
        Method & \multicolumn{2}{c}{Diversity}& \multicolumn{3}{c}{Counterfactual Search} \\
         & Diversity type  & Diversity criterion & Number of CF & Explicit & Single run\\
        \midrule

        LORE~\cite{guidotti2019lore} & Actions & Diverse leaves of a decision tree & algo & Yes & Yes\\
        
        Mahajan~\cite{mahajan2019} & Feature values & Stochasticity in the generation & user & No & Yes \\
        Russell~\cite{russell2019efficient} & Actions & Rerun $\&$ exclusion of the previous results & user$+$algo & Yes & No \\
        CADEX~\cite{moore2019explaining} & Feature values & Rerun $\&$ exclusion of the previous features used & user & No & No \\
        
        Ustun~\cite{Ustun2019} & Actions & Rerun \& exclusion of the previous results & user & Yes & No \\ 

        CERTIFAI\cite{sharma2020}  & Feature values & Exploration by sampling & user & No & Yes \\
        Tsirtsis~\cite{tsirtsis2020decisions} & Feature values & Partitioning of the data space & user & Yes & Yes \\
        
        MOC~\cite{Dandl_2020} & Feature values \& Criteria & Pareto front in objective space & algo & Yes & Yes \\
        
        MACE1~\cite{karimi2020model} & Features values & Rerun \& exclusion of the previous results & user & Yes & No \\
        
        DICE~\cite{Mothilal2020} & Feature values & Diversity term in the optimization & user & Yes & Yes \\
        
        DECE~\cite{cheng2020dece} & Feature values & Consideration of different constraints & user & Yes & Yes \\
        
        CRUDS~\cite{downs2020} & Feature values & Partitioning of the data space & user & Yes & Yes \\
        
        DiVE~\cite{Rodriguez_2021_ICCV} & Feature values & Diversity term in the optimization & user$+$algo & Yes & No \\
        OCEAN\cite{parmentier2021optimal} & Feature values & Rerun $\&$ exclusion of the previous results & user$+$algo & Yes & No \\
        MCCE~\cite{redelmeier2021mcce} & Feature values & Exploration by sampling & user & Yes & Yes \\
        OrdCE\cite{kanamori2021ordered} & Criteria & Pareto front in objective space & user & Yes & No \\
        
        MCS~\cite{yang2021model} & Features values & Exploration by sampling & user & No & No \\
        CSCF~\cite{naumann2021consequence} & Actions & Sequential approach to obtain CF with different sequences & user & Yes & Yes \\
        
        MIP-DIVERSE~\cite{mohammadi2021scaling} & Feature values & Rerun \& exclusion of the previous results & user & Yes & No \\
        
        Hada~\cite{hada2021exploring} & Feature values & Consideration of different constraints & user + algo & No & No \\
        
        Navas~\cite{navas2021optimal} & Feature values & Consideration of different constraints & user & Yes & Yes \\
        Samoilescu~\cite{samoilescu2021model} & Feature values & Partitioning of the data space & user & Yes & Yes \\
        
        Becker~\cite{becker2021step} & Actions & Diverse leaves of a decision tree & user + algo & No & Yes \\
        
        GeCo~\cite{Schleich2021GeCo} & Feature values & Exploration by sampling  & user & No & Yes\\
        FastAR~\cite{verma2021amortized} & Actions & Stochasticity in the generation & user & No & Yes\\
        Carreira~\cite{carreira2021counterfactual} & Feature values & Consideration of different constraints  & user & No & No \\
        EMC~\cite{yadav2021low} & Criteria & Initialisation of different cost functions & user & Yes & Yes \\
        
        $\delta$-CLUE~\cite{ley2022diverse} & Feature values & Diverse initialization for the optimization & user & No & No \\
        CARE1~\cite{rasouli2022care} & Criteria & Pareto front in objective space & user & Yes & Yes \\
        
        MACE2~\cite{Yang2022} & Feature values & Exploration by sampling & user & Yes & Yes \\
        
        Smyth~\cite{smyth2022few} & Feature values & k-NN model to delimit group of CF & user & Yes & Yes \\
        
        COPA~\cite{Bui2022} & Feature values & Optimization using gradient descent & user & Yes & Yes \\


        FRPD~\cite{nguyen2023feasible} & Feature values & Diversity term in the optimisation & user & Yes & Yes \\
        
    \end{tabular}}
    
    \caption{Summary of existing diverse counterfactual example (CF) generation methods, discussed in Section~\ref{sec:diverse-edl} that details the diversity type and criterion columns. The "Number of CF" column indicates whether the user can choose the number of desired counterfactual examples (user) or if the latter is automatically selected by the algorithm (algo). The "explicit" column indicates whether the diversity objective is explicitly included in the optimisation process or not. The "Single-step" column indicates whether the optimization applies a single-step (Yes) or an iterative procedure (No) to generate all the counterfactual examples.}
    \label{tab:diverse-survey}
\end{table*}

\subsubsection{Diversity in Criteria}
\label{sec:diverse-edl-criteria}
A first type of diversity definition depends on the quality criteria the counterfactual examples are required to optimize: it proposes to use different means to combine them, often relying on different aggregation operators instead of a single one. For example, Dandl et al.~\cite{Dandl_2020} and Rasouli et al~\cite{rasouli2022care} focus on generating multiple counterfactual examples, which all optimise the same criteria but perform different trade-offs between them. The generated counterfactual examples thus correspond to different positions on the Pareto front defined by the considered quality criteria, then chosen according to different strategies. To do so, Rasouli et al.~\cite{rasouli2022care} consider that a hierarchy among the different criteria is provided by the user. 
This avoids the risk induced by a trade-off operator that may lead to a solution that actually has a medium value for all considered criteria. Indeed, a user-defined hierarchy allows to select the criterion to be optimized first and those to be optimized later on. 

\subsubsection{Diversity in the Feature Space}
\label{sec:diverse-edl-feature}
A second type of diversity focuses on the relative position of the generated counterfactual examples in the feature space. According to this definition, diverse counterfactual examples lie far away from each other the feature space space. 
Here, the quality criteria are therefore not only used to evaluate the counterfactual examples individually, but also to analyze the relations between them. 
In this category, two strategies can be identified: the former induces diversity by first explicitly defining different constraints, and then generating counterfactual examples for each of them; the latter relies on defining diversity as a distance to be maximized between the generated counterfactual examples in the feature space. 

The constraints considered by the approaches relying on the first strategy can take various forms.
Most commonly~\cite{tsirtsis2020decisions,Rodriguez_2021_ICCV, navas2021optimal, carreira2021counterfactual}, they are defined to partition the feature space into several subspaces. Counterfactual examples are then generated in each of the identified subspaces, which allows to obtain explanations that are diverse as they belong to different areas of the space. For instance, \cite{tsirtsis2020decisions} relies on the user defining this partitioning along features of interest (e.g., age group) and generating one counterfactual per subspace. 

Approaches relying on the second strategy define diversity as a similarity or distance between counterfactual examples. This allows them to directly integrate diversity in the counterfactual generation. This can be done either by modifying the optimization problem by integrating a diversity criterion into the cost function, enabling them to a set of counterfactual examples at once: Equation~\ref{eq:generic-CF-problem} is then modified to:
\begin{equation}
    \{e^{*}_1, \dots, e^{*}_k\} = \argmin_{ \{e_1, \dots, e_k\} \subset \mathcal{E}} agg\left(\sum_{i=1}^k pen_x(e_i), \varphi(div(\{e_1, \dots, e_k\}))\right)
\label{eq:costFunctionDiverse}
\end{equation}
where $k$ denotes the number of desired counterfactual examples, $div$ is a function assessing their diversity, seen as a new quality criterion to be maximized, $\varphi$ a decreasing function and $agg$ an aggregation operator to combine the average quality of the counterfactual candidates and their diversity. The penalty function can obviously be combined with some of the additional criteria discussed in Section~\ref{sec:background-criteria}, such as sparsity, data or user contextualization. 

This diversity measure itself can take multiple forms. For example, several approaches focus on maximizing the diversity of features used in the final explanations \cite{russell2019efficient,bhatt2021divine,Rodriguez_2021_ICCV}, which can be translated as maximizing the $l_0$ distance between the proposed counterfactual examples. Other approaches, such as~\cite{Mothilal2020}, define diversity as the distance between the generated counterfactual examples (norm $l_1$, $l_2$, or both). As a result, the obtained counterfactual examples are thus distant from one another in the input space, and may use different features. 

Instead of defining a set of diverse counterfactual explanations as a solution to a single optimisation problem, other methods~\cite{russell2019efficient, hada2021exploring, mohammadi2021scaling} rely on an iterative process to generate the multiple explanations, a counterfactual example being generated at each step.  
To ensure that a new explanation is different from the previous ones, these approaches then consider constraints on the distance between the new explanation and the ones already generated.

In the case of non-binary classification, Ley et al.~\cite{ley2022diverse} propose to take into account, in addition to the feature space, the prediction space: they generate counterfactual examples associated with various classes among the ones different from the one predicted for the instance of interest. 


\subsubsection{Diversity in Actions}
\label{sec:diverse-edl-actions}
A counterfactual explanation by design suggests actions, as modifications to the instance of interest, that allow to get a different prediction. A third type of diversity aims at proposing explanations which need/use different actions. They are related to the diversity in terms of features discussed above, with a slightly different interpretation, more related to the notion of algorithmic recourse. Beside relying on the $l_0$ distance, this type of diversity can be achieved in more specific settings: 
Guidotti et al.~\cite{guidotti2019lore} and Becker et al.~\cite{becker2021step} use decision trees to generate explanations. Imposing the latter to be located in different leaves of the tree implies they follow different paths from the root to the leaves, and as a consequence rely on different actions. 

Instead of proposing explanations that modify different features, Russell et al.~\cite{russell2019efficient} propose explanations that go in different modification directions:
a first counterfactual example may recommend increasing the value of a given feature, whereas another one would recommend decreasing it. The induced actions are completely different. The same principle applies to the proposition of  Verma et al.~\cite{verma2021amortized}, that relies on performing successive small actions, in different directions, until obtaining the final explanation. 

\subsubsection{Optimisation-related dimensions}
Beside relying on different definitions of the notion of diversity, that apply at different levels, as discussed in the previous paragraphs, existing algorithms to generate multiple counterfactual examples also differ in the optimization procedure they apply, as we propose to discuss in this section. These comparison dimensions are summarised in the last three columns of Table~\ref{tab:diverse-survey}.

\paragraph{Explicit vs non-explicit diversity}
\label{sec:diverse-edl-explicit}
Independently from the discussions of the previous sections describing how diversity may being defined, another possibility to differentiating factor for methods is associated to how explicitly diversity is incorporated in the counterfactual generation. We thus make a distinction between \emph{explicit} and \emph{non-explicit} methods, and discuss them below.

\emph{Explicit} methods are characterized by the fact that they actively take into account diversity in the counterfactual counterfactual optimisation problem defined in Equation~\ref{eq:generic-CF-problem}. This can be achieved with all diversity definitions, in various ways.
A straightforward way is to include the notion of diversity directly in the cost function, so that the optimization of diversity is guaranteed between the counterfactuals like Mothilal et al.~\cite{Mothilal2020} or Dandl et al.~\cite{Dandl_2020}. The former propose to integrate it by defining the diversity of a set of solutions while the latter integrate it through the aggregation function that combines the different criteria. 
Other approaches focus on solving simultaneously different optimization problems with their own constraints, such as~\cite{carreira2021counterfactual, hada2021exploring}.
In this case, the information considered as input is not the same for each optimization problem, allowing to obtain explanations that answer different contexts or motivations and are thus diverse. 
Finally, some approaches propose to integrate diversity by excluding the explanations already generated from the set of possible solutions for later iterations, thus ensuring that the new explanations are different from the previous ones~\cite{parmentier2021optimal,russell2019efficient}. 
Thus, for explicit methods, there is a dedicated mechanism in the explanation generation process that ensures diversity (regardless of its definition).

On the other hand, non-explicit approaches are generally non-deterministic approaches, meaning that using the same approach twice in the same setting does not necessarily return the same explanation. Non-explicit counterfactual methods such as~\cite{mahajan2019,moore2019explaining,sharma2020} thus propose to generate diverse explanations by using the stochastic aspect of the generation process. 
Unfortunately, this means that the resulting diversity is not maximized nor guaranteed, as the set of generated examples may be close from one another for instance. 
Although some mechanisms are proposed to encourage diversity, such as in CERTIFAI~\cite{sharma2020} where the authors propose to modify the algorithm initialization, most of these remain unreliable when it comes to diversity.

\paragraph{Number of counterfactual examples returned}
\label{sec:diverse-edl-numberCF}
Another dimension related to diversity is the number of counterfactual examples that these methods allow to generate. 
Although most approaches propose to let the user set this number (e.g., among others~\cite{mahajan2019,Mothilal2020}), this choice may in some cases be limited by the method itself. For instance, for methods proposing to generate diverse explanations as examples belonging to different leaves of a tree~\cite{guidotti2019lore,becker2021step},  the number of counterfactual examples to be generated is bounded by the  number of leaves. However, despite the fact that the approaches that let the user set the number of desired counterfactual explanations seem to be less limited, they generally do not acknowledge that this number naturally often comes at odds with how diverse the explanations are. For many them (e.g. \cite{Ustun2019,Mothilal2020,sharma2020}), increasing this number will thus lead to the generation of redundant explanations.


\paragraph{One run vs. several runs}
\label{sec:diverse-edl-multipleruns}
To generate multiple explanations, two strategies exist: either all explanations are generated as solution of a single optimisation problem; or the optimisation problem only leads to the generation of a single counterfactual example, in which case several steps are required to generate iteratively additional explanations. Methods generating multiple explanations in 
a single step, referred to as "Single run" in Table~\ref{tab:diverse-survey}, often optimize a cost function applying to sets of candidates, as given in Equation~(\ref{eq:costFunctionDiverse}). The explanation diversity between explanations is then one of the criteria included in the considered cost function. Other methods focus on simultaneously exploring the data space in different directions (such as~\cite{tsirtsis2020decisions,guidotti2019lore}). 

On the other hand, other approaches~\cite{russell2019efficient,ley2022diverse} generate explanations iteratively, possibly using the explanations obtained in the previous steps to generate the new one.
Other approaches proposed by Carreira et al.~\cite{carreira2021counterfactual} or Samoilescu et al.~\cite{samoilescu2021model}  look at different constraints at each stage to address different contexts. For example, Carreira et al. increase the number of considered constraints at each stage. Thus, at each step the algorithm is defined with different constraints. The studied optimization problem is then different for each counterfactual example.

\section{Discussion and Research Challenges}
\label{sec:discussion}

The richness of the diverse counterfactual approaches presented in the previous section underlines once again the importance of properly motivating the choice of the explanability objectives, including the diversity. In this section, we propose a discussion on the link between the diversity notions identified, and how they may better help with addressing user needs.


\subsection{Diversity as a Way to Match Unknown User Needs}
Section~\ref{sec:diverse-motivations} reminds one of the strongest arguments supporting the use of multiple counterfactual explanations: their ability to help match unobserved user needs. An implicit assumption for this purpose is the explanations had to be \emph{diverse}~\cite{sullivan2022explanation}, leading to concurrent definitions of diversity, described in the previous section. Yet, we argue that all diversity definitions do not all unequivocally fulfill this promise.

Intuitively, \emph{Diversity in Criteria}, described in Section~\ref{sec:diverse-edl-criteria}, directly addresses this objective: by proposing multiple ways of combining different quality criteria, they allow the user to choose his/her own order of preference between the explanations' properties. For example, asking a user to specify the minimum level of sparsity of the explanation he wants for a problem might be complicated for him. Offering explanations with different levels of sparsity and penalty could help him to understand the trade-off between these two notions in the explained prediction and to select his preferred aggregation. Yet, this requires the user to be able to understand, if not the property captured, at the very least that the proposed diverse explanations vary along these criteria. This therefore questions the use of the notion of diversity in criteria for data-contextualization criteria (cf. Section~\ref{sec:background-criteria}), which are arguably (i) much harder for a user to understand and (ii) not directly visible when looking at presented counterfactual explanations, at least without additional context. 
User-contextualization criteria, on the other hand, do not suffer from this issue. By nature, although the numerical quantification of these properties can be challenged, it is expected that the user directly understands the differences. 
This leads us to believe that to be relevant, the diversity of a set of counterfactual explanations should be observable. This naturally questions the utility of approaches integrating a \emph{non-explicit} diversity, as they do not guarantee the resolution of the problem.

By proposing explanations that provide a set of actions that vary in terms of targeted features, approaches integrating \emph{Diversity in Actions} fulfill this objective of observable differences. The user is provided several alternative recourses, among which he/she may choose their preferred according to their internal unobserved preferences. On the other hand, \emph{Diversity in Feature values} do not explicitly address a formulated user need. Apart from being a proxy for Diversity in Actions when user-contextualization criteria are not available, they thus do not seem to answer user expectations, despite being the most represented type of approach (see Table~\ref{tab:diverse-survey}).  A notable exception would be in the context of model debugging, where such diversity might be providing relevant information about the local decision boundary and feature importance weights.


\subsection{Diversity Beyond Counterfactual Explanations}

On the contrary to most of the presented quality criteria, the penalty of the explanations is rarely included as a property to balance. Due to the usual formulation of counterfactual explanations, penalty is indeed generally seen as a criterion to minimize, or eventually that it is possible to sacrifice a little bit, only when necessary, to satisfy other properties. Few works thus explore the possibility of explicitly combining local counterfactual explanations with more global ones (a concept proposed for instance by~\cite{rawal2020beyond}). 
Yet, several works have showcased in applied contexts the benefits of combining local and global explanations on interpretability. 
Such works include for instance~\cite{collaris2018instance}, mixing local and global feature importances to explain fraud detection models, or~\cite{bove2022contextualization}, proposing local explanations with global contextualization to help customers understand insurance pricing. 
Integrating penalty as one criteria to balance among others for counterfactual explanations thus represents an interesting perspective. 
As illustrated by the works mentioned above, this seems strongly related to another notion of diversity that could be formulated, also out of the scope of this paper, which is the diversity in explanations forms.

\section{Conclusion}
\label{sec:conclusion}

In this work, we show how the combination of the quality criteria considered to generate counterfactual explanations is not trivial and may inaccurately match the needs of the user, which are often unobserved and only implicitly defined. We argue this provides new arguments in favour of approaches that propose 
diverse counterfactual explanation as a solution to this issue, letting the users choose their preferred explanation.
As discussed in the paper, various definitions of diversity have been proposed in the literature, the conducted analysis shows that they do not equally help addressing user needs. 
Future works will aim at conducting an empirical study to complement this analysis, in a full human-in-the-loop paradigm, so as to examine thoroughly the impacts these diversity definitions can have on users and the respective cases where they might appear more appropriate. This direction of research also calls for new tools to represent and model user needs as well as user knowledge, so as to establish a correspondence with the most relevant approaches. 


\begin{acks}
This research was supported by TRAIL, a joint laboratory between SORBONNE UNIVERSITE/CNRS (LIP6) and AXA.
\end{acks}

\bibliographystyle{ACM-Reference-Format}
\bibliography{biblio}
\end{document}